\documentclass[conference]{IEEEtran}
\IEEEoverridecommandlockouts    

\usepackage{amsmath} 
\usepackage{amssymb}  
\usepackage{siunitx}
\usepackage{multirow}
\usepackage{xspace}

\usepackage{tikz}
\usepackage{pgfplots}
\usepackage{tikzscale}
\usepackage{marvosym}
\usepackage{balance}

\usepackage{graphicx}

\usepackage{algorithm}
\usepackage{algorithmicx}
\usepackage[noend]{algpseudocode}

\usepackage[caption=false]{subfig}

\usepackage{jsonparse}

\makeatletter

\pgfplotsset{compat=1.18}
\usepackage{pgfplotstable}
\usepackage{pgfplots}
\usetikzlibrary{external}
\usetikzlibrary{matrix, decorations.pathreplacing, calc, positioning, fit, fadings, shapes.arrows, arrows.meta, shapes.geometric, backgrounds}
\usetikzlibrary{shadings}
\usetikzlibrary{fadings,decorations.pathreplacing}
\usetikzlibrary{shapes.arrows}
\usetikzlibrary{math}
\usetikzlibrary{automata}
\usetikzlibrary{positioning}
\usepgfplotslibrary{fillbetween}
\usepgfplotslibrary{statistics}
\definecolor{custom-blue}{HTML}{1a80bb}
\definecolor{custom-red}{HTML}{a00000}
\definecolor{custom-yellow}{HTML}{f2c45f}
\definecolor{muted-gray}{HTML}{808080}
\definecolor{muted-gold}{HTML}{f0c571}
\definecolor{muted-teal}{HTML}{59a89c}
\definecolor{muted-blue}{HTML}{0b81a2}
\definecolor{muted-red}{HTML}{e25759}
\definecolor{muted-darkred}{HTML}{9d2c00}
\definecolor{muted-purple}{HTML}{7e4794}
\definecolor{muted-green}{HTML}{36b700}
\definecolor{muted-darkblue}{HTML}{082a54}
\definecolor{grid-green}{HTML}{67a05f}
\definecolor{grid-blue}{HTML}{0000ff}
\definecolor{mec-red}{HTML}{dc0000}

\definecolor{dark-red}{HTML}{c31e23}
\definecolor{light-red}{HTML}{ff5a5e}
\definecolor{med-brown}{HTML}{ea801c}
\definecolor{light-brown}{HTML}{f0b077}

\tikzstyle{arrow} = [line width=0.7pt,->,>=stealth]
\tikzstyle{doublearrow} = [thick,<->,>=stealth]
\tikzstyle{box} = [draw, rectangle, minimum width=1.9cm, minimum height=0.9cm, font=\footnotesize]

\def \minwidth {1.2cm}
\tikzstyle{rounded box} = [draw, rectangle, rounded corners, minimum height=.7cm, minimum width=\minwidth, font=\scriptsize, line width=0.7pt]

\makeatletter
\newcommand\footnoteref[1]{\protected@xdef\@thefnmark{\ref{#1}}\@footnotemark}
\makeatother

\newcommand{\ros}{ROS~2\xspace}

\newcommand\HUGE{\@setfontsize\Huge{40}{50}}
\makeatother
\title{\LARGE \bf
Extending Responsibility-Sensitive Safety for the Assessment of Offloaded Autonomous Driving Services 
}

\author{Robin Dehler, Aryan Thakur, and Michael Buchholz
\thanks{All authors are with the Institute of Measurement, Control and Microtechnology, Ulm University, Albert-Einstein-Allee 41, 89081 Ulm, Germany {\tt\footnotesize \{firstname\}.\{lastname\}@uni-ulm.de}}%
}
\newcommand\copyrighttext{\footnotesize \textcopyright~2026 IEEE. Personal use of this material is permitted. Permission from IEEE must be obtained for all other uses, in any current or future media, including reprinting/republishing this material for advertising or promotional purposes, creating new collective works, for resale or redistribution to servers or lists, or reuse of any copyrighted component of this work in other works.%
}
\newcommand\copyrightnotice[1]{%
    \begin{tikzpicture}[remember picture,overlay]%
     \node[%
        anchor=south, %
        yshift=#1pt%
    ] at (current page.south)%
     {\fbox{\parbox{\dimexpr\textwidth-\fboxsep-\fboxrule\relax}{\copyrighttext}}};%
     \end{tikzpicture}%
}

\begin{document}
	
\maketitle
\thispagestyle{empty}
\pagestyle{empty}

\begin{abstract}

Safety is a fundamental requirement in the development of autonomous driving (AD) systems.
While function offloading has demonstrated significant benefits in terms of computational efficiency and energy consumption, its application to safety-critical AD functionality introduces new challenges.
In particular, offloaded service compositions incur increased and variable response times due to wireless vehicle-to-everything (V2X) communication, which directly affects the vehicle’s reaction time and thus its safety guarantees.
In this paper, we address this challenge by extending the definitions of Responsibility-Sensitive Safety (RSS) to explicitly account for different response times of local and offloaded AD service compositions.
Based on this extension, we propose an integration into function offloading, using the RSS safety constraints for offloading decision-making and fallback mechanisms.
Off\-loaded service compositions are only permitted if the current traffic situation remains safe under the corresponding end-to-end response time.
If this condition is violated, the system performs a controlled fallback to local execution.
Furthermore, we introduce an enhanced fallback strategy that includes a warm-standby phase for offloaded services, enabling faster and safer transitions from offloaded to local services.
The proposed approach is integrated into our AD stack and evaluated in both simulation and the real world.
Experimental results demonstrate that the proposed method improves safety compared to state-of-the-art function offloading and safety frameworks, while preserving the benefits of distributed computation when safety conditions allow.

\end{abstract}
\copyrightnotice{5}
\section{Introduction}
Many of the challenges in autonomous driving (AD) arise from the limited computational and energy resources available in autonomous vehicles.
Recent advances in vehicle-to-everything (V2X) communication enable the application of function offloading techniques to mitigate these limitations by leveraging external computational resources for connected autonomous vehicles (CAVs).
A call for adopting function offloading is evident, with performance characteristics analyzed in real-world testing of 5G networks~\cite{Dettinger-2025}.
In function offloading, computational tasks are distributed among heterogeneous processing units, including on-board systems, edge infrastructure, and cloud servers~\cite{saeik-21}.
However, function offloading alone does not specify mechanisms for the standardized description, coordination, and runtime management of distributed tasks.
Service-oriented architectures (SOAs) address this limitation by encapsulating functionality as services with well-defined interfaces, thereby enabling flexible deployment and dynamic reconfiguration of computational tasks~\cite{kampmann-19}.
Within this architectural framework, we have presented a service orchestrator that defines the service composition, including the execution location of individual services, in our previous work~\cite{dehler-25}.

When using offloaded services, however, other challenges arise.
The consideration of quality of service (QoS) and data safety and security becomes critical.
Consequently, the assessment of offloaded services is an important component for safely managing the AD task and preventing accidents.

Responsibility-Sensitive Safety (RSS) mathematically defines whether a traffic situation is considered safe or dangerous~\cite{schwartz-17}.
This analysis is based on a set of rules that compare the state of an ego vehicle with the states of other vehicles.
These rules are inherently dependent on the response time, i.e., the time required to compute control variables from sensor input.
Considering offloaded services, the response time depends not only on the execution time of each service to generate control output but also on the time required for V2X communication, i.e., exchanging relevant data between different computing units.
Depending on the service composition, the time for V2X communication is a relevant factor in the overall end-to-end response time, especially when large amounts of data need to be transmitted, e.g., raw sensor data or grid maps~\cite{dehler-26-1}.
\begin{figure}[t]
    \centering
    \includegraphics[width=.99\linewidth]{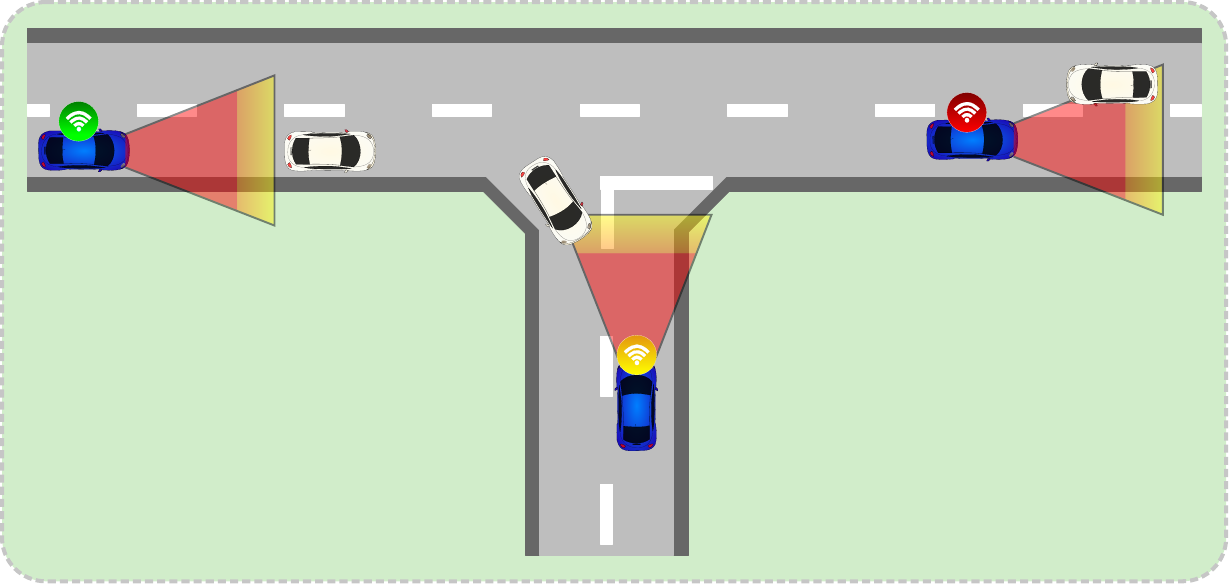}
    \caption{Simplified illustration of three different RSS scenarios, including one for the proposed RSS extension for CAVs. The blue vehicles are the considered CAVs. While the left scenario depicts a safe situation and the right scenario a dangerous one, the scenario in the middle depicts an RSS scenario for the proposed extension, i.e., a guarded situation.}
    \label{fig:teaser}
\end{figure}

In this paper, we extend RSS to account for different response times depending on the service composition of distributed SOAs.
The proposed extension introduces a novel third RSS state to classify situations.
This new RSS state, denoted as guarded, enables the preparation of a fallback during function offloading, creating a safer switch back to a local service composition.
We integrate the RSS extension into our previously proposed service-oriented function offloading framework (SOFOF)~\cite{dehler-25} to assess the situation during function offloading and, if required, adjust the service composition for a safer execution of the AD task.


The proposed extension is illustrated in Fig.~\ref{fig:teaser}, showcasing the three different RSS states.
In the top-left panel, a safe scenario (green antenna icon) is shown, where the blue CAV maintains a sufficient distance from the vehicle in front.
The situation at the top-right is dangerous (red antenna icon), as the other CAV is within the red envelope of the blue CAV.\footnote{The choice of using envelopes for the safety distance is a simplified version for illustration purposes of the true RSS checks.}
A guarded state (yellow antenna icon) is illustrated in the middle, where the white vehicle is not yet in the dangerous red envelope but is still in the yellow guarded state, if offloaded services are used.

This assessment is included as an additional safety criterion for the function offloading decision-making.
In a nutshell, function offloading shall only be used when the situation meets the safety requirements defined by our proposed extended RSS model.
We evaluate our approach in simulation by comparing it with different baseline function offloading frameworks, highlighting the differences in offloading decision-making between the extended RSS-aware CAV and conventional approaches that do not incorporate RSS-based safety constraints.
Lastly, we have also deployed the proposed method in our real-world test CAV, providing proof of concept for real-world traffic and network scenarios.

The remainder of this paper is organized as follows:
\begin{itemize}
    \item We present the background of RSS, SOA, and function offloading, including state-of-the-art methods in Sec.~\ref{ch:background}.
    \item We extend RSS to support distributed service compositions and present our method for integrating it into SOFOF for safe offloading decision-making and for fallback preparation in Sec.~\ref{ch:method}.
    \item We analyze the performance of our approach in simulation, comparing it to state-of-the-art function offloading frameworks, and in a real-world test setup in Sec.~\ref{ch:evaluation}.    
\end{itemize}

\section{Background}\label{ch:background}
In this section, the relevant background for RSS and service-oriented function offloading is shortly summarized.

\subsection{Responsibility-Sensitive Safety}
RSS is a well-known approach for analyzing the safety of an autonomous vehicle.
In RSS, traffic scenarios are analyzed for an ego vehicle based on its own state relative to the states of surrounding traffic participants.
Safety is considered similarly to how we humans should consider it, i.e., through the \textit{duty of care} from \textit{Tort law}~\cite{schwartz-17}.
Based on this, RSS formalizes safety mathematically, accounting for safety distances based on vehicle states, response times, and vehicle-specific parameters, e.g., maximum (braking) acceleration. 
A scenario is classified as RSS-safe or RSS-dangerous if a danger threshold time that is proportional to the safety distance is exceeded\footnote{In the following, we use the terminology \textit{RSS-safe} or \textit{RSS-dangerous} to specifically indicate the RSS states.}.
Then, a proper response, depending on the cause of the RSS-dangerous situation, should be triggered.
In RSS, the rules always apply to all vehicles.
However, since we assume we do not control the other vehicles, we consider RSS from the perspective of an ego vehicle.
For the scope of this paper, we shortly introduce the basic ideas of RSS and its formulations derived from~\cite{schwartz-17}.

\subsubsection{Safety distances}
The safety distances for RSS are defined longitudinally and laterally, considering a lane-based coordinate system.
Given the states of an ego vehicle $\nu_\text{e}$ and a considered other vehicle $\nu_\text{o}$, the longitudinal safety distance can be defined to
\begin{align}\label{eq:londist}
    &d_\text{min}^\text{lon}= \\
    &\begin{cases}
    \left[v_\text{e}\rho+\frac{a_\text{max,a}^\text{lon}\rho^2}{2}+\frac{(v_\text{e}+\rho a_\text{max,a}^\text{lon})^2}{2a_\text{min,b}}-\frac{v_\text{o}^2}{2a_\text{max,b}^\text{lon}}\right]_+ &\text{, same} \\
    \frac{v_\text{e}+v_{\text{e},\rho}}{2}\rho+\frac{v_{\text{e},\rho}^2}{2a_\text{min,b}^\text{lon}}+\frac{|v_\text{o}|+v_{\text{o},\rho}}{2}\rho+\frac{v_{\text{o},\rho}^2}{2a_\text{min,b}^\text{lon}} &\text{, opposite} \\
    \end{cases}. \nonumber
\end{align}

The upper case in Eq.~\eqref{eq:londist} applies if $\nu_\text{e}$ and $\nu_\text{o}$ drive in the same direction, and---without loss of generality---if $\nu_\text{e}$ is behind $\nu_\text{o}$.
The values $v_\text{e}$, $v_\text{o}$ are the longitudinal speeds of $\nu_\text{e}$ and $\nu_\text{o}$, respectively, $a_\text{max,a}^\text{lon}$ is the maximum longitudinal acceleration, $a_\text{min,b}^\text{lon}$ the minimum braking acceleration, $a_\text{max,b}^\text{lon}$ the maximum braking acceleration, and $\rho$ the response time.\footnote{For simplicity in this definition, we considered the minimum and maximum accelerations to be the same for both $\nu_\text{e}$ and $\nu_\text{o}$}
The notation $\left[d\right]_+ = \max\{0,d\}$ was taken from~\cite{schwartz-17}.
The case below applies if the vehicles drive in opposite directions, meaning $v_\text{o}<0$ and $v_\text{e}\geq0$, with $v_{\text{e},\rho} = v_\text{e}+\rho a_\text{max,a}$, and $v_{\text{o},\rho} = |v_\text{o}|+\rho a_\text{max,a}$.

The lateral safety distance can be defined similarly to
\begin{align}\label{eq:latdist}
    &d_\text{min}^\text{lat}= \\
    &\mu+\left[\frac{v_\text{e}+v_{\text{e},\rho}}{2}\rho+\frac{v_{\text{e},\rho}^2}{2a_\text{min,b}^\text{lat}}-(\frac{v_\text{o}+v_{\text{o},\rho}}{2}\rho-\frac{v_{\text{o},\rho}^2}{2a_\text{min,b}^\text{lat}})\right]_+, \nonumber
\end{align}
now with lateral accelerations $a^\text{lat}$ and lateral velocities $v_\text{e}$, $v_\text{o}$, and a minimal lateral velocity $\mu$.
In Eq.~\eqref{eq:latdist}, $v_\text{e}$ and $v_\text{o}$ are now lateral speeds.

\subsubsection{Dangerous situation}
A situation between two vehicles is considered as RSS-dangerous if both the longitudinal and lateral safety distances, i.e., $d_\text{min}^\text{lon}$ and $d_\text{min}^\text{lat}$, are violated. 
The danger threshold times $t_b^\text{lon}$ and $t_b^\text{lat}$ define the earliest time steps at which the situation is considered RSS-dangerous longitudinally or laterally, respectively.

\subsubsection{Proper response}
If a situation is deemed RSS-dangerous, a proper response needs to be triggered. This proper response is either longitudinal, if $t_b^\text{lon}\geq t_b^\text{lat}$, or otherwise lateral.
The longitudinal proper response for $\nu_\text{e}$ is to decelerate the vehicle with more than the minimally considered braking acceleration, i.e., $a_\text{min,b}$.
In the case of the same direction, this is applied until a safe longitudinal situation is reached; in the case of the opposite direction, it is applied until reaching a full stop.
The lateral proper response is to apply lateral acceleration with at most $-a_\text{min,b}^\text{lat}$, until reaching a $\mu$-lateral velocity of $0$.

\subsubsection{More complex scenarios}
The longitudinal and lateral safety distances are defined for a lane-based coordinate system based on the lane on which $\nu_{e}$ drives.
Consequently, the states of the other vehicles are mapped to the ego lane.
This is insufficient for more complex scenarios, e.g., intersections or roundabouts.
For more complex scenarios, the paths on which the other vehicles drive need to be considered.
Then, given an intersection point with the lane of the ego vehicle, the formulations for longitudinal and lateral safety distances, and the proper response can be applied again.
For a more detailed explanation of RSS, we refer to~\cite{schwartz-17}.

\subsection{Service-Oriented Function Offloading}
In service-oriented function offloading for CAVs, the definitions of automotive service-oriented architectures (ASOA)~\cite{kampmann-19} are used to apply function offloading to AD services.
Then, the set of possible service compositions for achieving safe AD can be extended to include available remote services.
These remote services may be offered by, e.g., an edge or cloud server, or other CAVs with available computing capabilities.

In~\cite{dehler-25}, we have proposed SOFOF.
The framework is split into a service requester part used to request offloaded remote services, and a service provider part that runs on the instance providing additional remote services.
In SOFOF, service orchestration, i.e., the management of the selected service composition, is handled jointly by both instances, with the service provider informing the service requester about available remote services so the requester can select an appropriate service composition for the AD task.
For successful function offloading, an offloading decision-making algorithm can determine whether offloading is worthwhile in a given situation.
In~\cite{dehler-25}, we proposed a simple location-based offloading decision-making algorithm, where offloading is possible if a service is available in a given area, e.g., considering network availability. 
A fallback to only local services may be triggered if a QoS violation is detected.

\begin{figure}[t]
    \centering
    \input{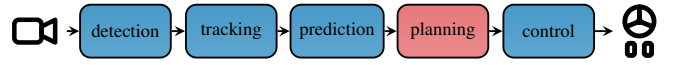}
    \caption{Simplified modular service configuration for an offloaded trajectory planning service. Local services are indicated in blue, while the remote planning service is in red. } 
    \label{fig:service-comp}
\end{figure}

In this paper, similarly to~\cite{dehler-25}, we exemplarily consider the use case of offloading the service trajectory planning.
Offloading of other services, e.g., tracking or detection, is also possible but excluded from our evaluation.
A simplified service composition that includes the offloaded planning service is shown in Fig.~\ref{fig:service-comp}.
When applying function offloading for trajectory planning, relevant data must be exchanged between the CAV and the edge server.
This data mainly includes the tracks, possibly including predictions, generated from the detections sensed by the CAV's sensors, the path the CAV wants to follow, and the vehicle state.
The sending of the data, including adherence with QoS constraints, is inherently integrated into the present SOFOF.
Our contribution enhances SOFOF by incorporating the RSS formulation into the offloading decision-making and fallback mechanism. 

\subsection{State of the Art}
Function offloading has been approached from a wide range of optimization objectives, e.g., by considering network conditions and energy efficiency~\cite{saeik-21, acheampong-22}, and reinforcement learning-based methods address resource allocation, minimizing latency, and energy consumption~\cite{Hortelano-2023}.
To the best of our knowledge, existing approaches do not condition the offloading decision on whether the ego vehicle is currently in a safe situation.

However, safety considerations are crucial for the development of holistic AD functions.
Consequently, safety-related V2X approaches for AD have already attracted interest~\cite{liu-19}.
These approaches can be categorized into those that use V2X to improve safety and those that safeguard V2X technologies.
Safety improvements enabled by V2X technology result from enhanced environment models with track-to-track fusion~\cite {cress-2024, buchholz-22} or cooperative approaches~\cite{hult-2016, haefner-22, klimke-26}. 
Simultaneously, V2X itself must be safe and secure, with cybersecurity being closely related.
A comprehensive survey on this topic is presented in~\cite{sedar-23}. 

In~\cite{dehler-26}, we presented a method for assessing offloaded AD services.
The proposed MUFASA framework can be used to validate data from offloaded services, such as trajectory planning and tracking.
MUFASA extends native SOFOF with an improved fallback mechanism if data validation fails during active function offloading.
Regarding safety constraints specifically, a cooperative driving framework that extends the RSS rules to handle trajectories for safety considerations is presented in~\cite{khayatian-21}.
The same authors extended their work using the control barrier approach to enforce safety constraints~\cite{khayatian-24}.
\section{RSS-Aware Function Offloading}\label{ch:method}
In this paper, we propose an extension of RSS for integration into SOFOF.
Our proposal is based on the following principle: 
Perform function offloading only when a scenario satisfies the extended RSS safety constraints.
In the following, we present the method in more detail.

\subsection{Method}

The classification of scenarios into RSS-safe or RSS-dangerous is mainly dependent on the minimum safety distance $d_\text{min}^\text{lon/lat}$ to other vehicles.
This safety distance depends on the vehicle states and the response time $\rho$ of the AD stack.
Depending on the current service composition, the response time $\rho$ differs strongly.
This is especially true when using offloaded services, where data is typically exchanged over a wireless transmission medium between different computing units.
Our approach of integrating RSS to improve the safety and scalability of function offloading for CAVs leverages different response times.
The idea is to consider different stages with different response times for local execution and offloading.

The protocol is summarized in Alg.~\ref{alg:method}:
The input into the protocol is the response times $\rho_\text{local}$, $\rho_\text{off}$, and $\rho_\text{switch}$, the state of the ego vehicle $s_\text{ego}$, as well as a list including the states of all other vehicles $[s_\text{other}]$.
Note that the states of the other vehicles might be present as tracks.
The algorithm's output is a boolean indicating whether offloading is possible based on the RSS classification of the current state: true if the state is RSS-safe, false if RSS-dangerous.

Before choosing an offloaded service composition (line~1), the RSS rules with the local response time $\rho_\text{local}$ need to return RSS-safe for all vehicles (lines~2-5).
If an offloaded service composition is chosen (line~6), the RSS rules are applied with the offloading response time $\rho_\text{off}$ (line~8).
If the RSS rules return RSS-dangerous, however, we do not directly stop offloading (line~9); instead, the RSS rules are repeated with the sum of $\rho_\text{local}$ and $\rho_\text{switch}$, i.e., the time it takes to switch back to a local service composition (line~10).
If it is RSS-safe, then the offloaded service composition can still be active, but a fallback is prepared (lines~13-14) to ensure a safe transition to a local service composition if this check eventually returns RSS-dangerous (lines~11-12).
If offloading is stopped due to an RSS-dangerous situation, the entire process restarts from the beginning.

\begin{algorithm}[t]
    \caption{RSS-Aware Function Offloading Protocol}
    \vspace{5pt}
    \label{alg:method}
    \hspace*{\algorithmicindent} \textbf{Input:} $\rho_\text{local}$, $\rho_\text{off}$, $\rho_\text{switch}$, state $s_\text{ego}$, states $[s_\text{other}]$\\
    \hspace*{\algorithmicindent} \textbf{Output:} Bool for offloading possibility
    \vspace{2pt}
    \begin{algorithmic}[1]
    \If{offloading inactive}
        \For{$s_\text{o}$ in $[s_\text{other}]$}
            \State state$_\text{RSS,local}$ = RSS($\rho_\text{local}$, $s_\text{ego}$, $s_\text{o}$)
            \If{state$_\text{RSS,local}$ \textbf{is} dangerous}
                \State Trigger proper response
                \State \textbf{return} false
            \EndIf
        \EndFor
    \Else
        \For{$s_\text{o}$ in $[s_\text{other}]$}
            \State state$_\text{RSS, off}$ = RSS($\rho_\text{off}$, $s_\text{ego}$, $s_\text{o}$)
            \If{state$_\text{RSS, off}$ \textbf{is} dangerous}
                \State state$_\text{RSS, switch}$ = RSS($\rho_\text{local} + \rho_\text{switch}$, $s_\text{ego}$, $s_\text{o}$)
                \If{state$_\text{RSS, switch}$ \textbf{is} dangerous}
                    \State Switch back to local execution
                    \State Trigger proper response
                    \State \textbf{return} false
                \Else
                    \State Prepare fallback
                \EndIf
            \EndIf
        \EndFor
    \EndIf
    \State \textbf{return} true
    \end{algorithmic}
    \vspace{2pt}
\end{algorithm}

The presented RSS-aware function offloading protocol ensures that a CAV always resolves RSS-dangerous scenarios using a local service composition.
Simultaneously, the time when offloading is possible is the time when the scenario is RSS-safe, either with $\rho_\text{local}$, if offloading is active for the first time on the respective server, or with $\rho_\text{off}$, if the check already failed once.
Including the fallback preparation mechanism, this method thus safeguards offloaded service compositions from potential accidents.
Consequently, we denote the state where the situation is RSS-dangerous with the response time $\rho_\text{off}$, but RSS-safe with $\rho_\text{local} + \rho_\text{switch}$, as RSS-guarded.

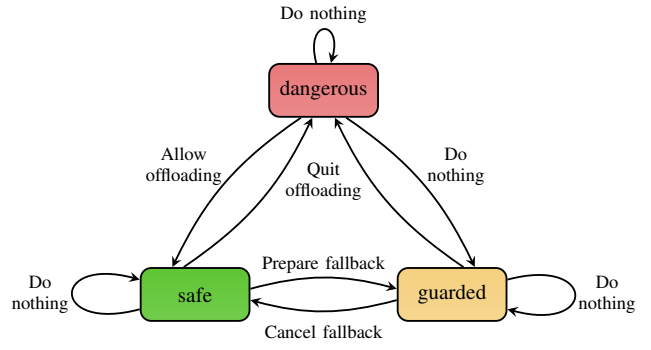
\begin{figure}[t]
    \centering
    \begin{tikzpicture}
\tikzstyle{statemachine} = [rounded box, minimum height=.7cm, minimum width=1.2*\minwidth, font=\footnotesize]

\node (d) [statemachine, align=center, top color=muted-red!80, bottom color=muted-red!70] {dangerous};
\node (s) [statemachine, align=center, below of=d, xshift=-1.7cm, yshift=-1.7cm, top color=muted-green!80, bottom color=muted-green!70] {safe};
\node (g) [statemachine, align=center, below of=d, xshift=1.7cm,, yshift=-1.7cm, top color=custom-yellow!80, bottom color=custom-yellow!70] {guarded};

\draw [arrow,>=stealth] ($(d.south) + (-0.3cm, 0cm)$) to [bend right=15] ($(s.north) + (-0.3cm, 0)$);
\draw [arrow,>=stealth] ($(s.north) + (-0.15cm, 0)$) to [bend right=15] ($(d.south) + (-0.15cm, 0)$);
\node (ds) [below of=d, xshift=-1.85cm, yshift=0cm, font=\scriptsize, align=center] {Allow\\offloading};

\draw [arrow,>=stealth] ($(d.south) + (0.3cm, 0cm)$) to [bend left=15] ($(g.north) + (0.3cm, 0)$);
\draw [arrow,>=stealth] ($(g.north) + (0.15cm, 0)$) to [bend left=15] ($(d.south) + (0.15cm, 0)$);
\node (dg) [below of=d, xshift=1.75cm, yshift=0cm, font=\scriptsize, align=center] {Do\\nothing};
\node (gd) [below of=d, yshift=-.2cm, font=\scriptsize, align=center] {Quit\\offloading};

\draw [arrow,>=stealth] ($(s.east) + (0, 0.075cm)$) to [bend left=15] ($(g.west) + (0cm, 0.075cm)$);
\draw [arrow,>=stealth] ($(g.west) + (0cm, -0.075cm)$) to [bend left=15] ($(s.east) + (0, -0.075cm)$);
\node (sg) [below of=d, yshift=-1.3cm, font=\scriptsize] {Prepare fallback};
\node (gs) [below of=d, yshift=-2.2cm, font=\scriptsize] {Cancel fallback};

\draw [arrow,>=stealth] (d) to [loop above] (d);
\node (dd) [above of=d, font=\scriptsize] {Do nothing};
\draw [arrow,>=stealth] (s) to [loop left] (s);
\node (ss) [left of=s, xshift=-1.05cm, font=\scriptsize, align=center] {Do\\nothing};
\draw [arrow,>=stealth] (g) to [loop right] (g);
\node (ss) [right of=g, xshift=1.05cm, font=\scriptsize, align=center] {Do\\nothing};

\end{tikzpicture}

    \caption{State machine including the typical RSS states safe and dangerous, as well as the extension for offloaded services, i.e., the guarded state. At each state transition, the consequences for function offloading are indicated.} 
    \label{fig:state-machine}
\end{figure}

In Fig.~\ref{fig:state-machine}, the three RSS states for offloaded service compositions, i.e., RSS-safe, RSS-guarded, RSS-dangerous, including possible state transitions, are shown.
The actions for function offloading that correspond with the state transitions are labeled accordingly.
To summarize, genuine offloading is allowed only in the RSS-safe state; in the RSS-guarded state, a fallback preparation is required; and in the RSS-dangerous state, offloading is disabled completely.

\subsection{Fallback Preparation}
Our method proposal includes a fallback preparation to directly handle an RSS-dangerous scenario that occurs after switching back to a local service composition from the RSS-guarded state.
When an offloaded service composition is present, we set the equivalent local services to be inactive.
Since the local services are only inactive and not shut down, the switch-back time to activate the services is comparably small, i.e., $\rho_\text{switch} << \rho_\text{local} < \rho_\text{off}$.

Our fallback consists of two components to ensure an immediate handling of the potentially dangerous situation.
First, when a transition to the RSS-guarded state occurs, we already set specific parameters accordingly, so that after the switch back, the newly activated local service composition immediately executes the proper response.
Second, we use a warm-standby mechanism for the offloaded services, based on the formalism of standby redundant systems. 

Considering the complete deactivation as the cold-standby phase and the normal active configuration of the service as hot-standby, the warm-standby phase for each service must be defined.
For the considered trajectory planning service, we define the warm-standby phase as the period when the node is activated but runs at a lower update frequency.
For other services, e.g., tracking, a similar idea can be employed.

\subsection{Analysis}
Here, we critically analyze the integration of the proposed RSS extension into offloading decision-making.
The integration of safety-based offloading decision-making is expected to reduce the overall time available for offloading.
Consequently, the efficiency benefits of function offloading are reduced.
As is typical when considering efficiency, our proposed method depicts a trade-off for safety.
We argue that when safety is addressed, this trade-off is worthwhile.

The same applies to fallback preparation when applying the warm-standby phase to the offloaded AD services.
This approach compromises efficiency for safety, since a warm-standby is more efficient than fully activating everything; however, it is still less efficient than keeping it inactive until the service is actually used.

\subsection{Implementation}
We have implemented the RSS rules as part of our modular AD stack.
Each module of the AD stack is implemented as a \ros lifecycle node.
A subset of the modules is seen in Fig.~\ref{fig:service-comp}.
The full set of modules, however, includes other modules, e.g., SOFOF.
In the context of SOA, the SOFOF module also extends our \ros nodes to be handled by the integrated service orchestrator, including distributed services.

Since the modules of the AD stack are highly interdependent, calculating the response times is not trivial, especially not online.
Thus, we obtain response times offline.
For this, we create and spawn a dummy object in front of the vehicle and measure the time difference between the object's spawn and the vehicle's reaction to it.
An analysis of this method is part of our evaluation.

To apply the RSS rules, the implemented RSS node subscribes to \ros topics, retrieving the required data, e.g., the ego vehicle state and tracks.
The minimum safety distances can then be calculated using the measured response times for the local and offloaded service compositions and the switch-back time.
Vehicle-to-lane mapping is performed using the Frenet transformation~\cite{doCarmo-76}.

Communication with the other nodes is equivalently done using the \ros middleware.
Depending on the RSS state or RSS state transition, the SOFOF node and offloaded services are notified and adjusted.
\section{Evaluation}\label{ch:evaluation}
In this section, after describing our setup, we evaluate our method in simulation and a real-world deployment.

\subsection{Experimental Setup}
We first analyze the performance of our method using a software-in-the-loop (SIL) simulation.
In the simulation, we spawn CAVs and several other simulated vehicles.
The simulation runs on an AMD Ryzen Threadripper 3990X CPU with 128 GB RAM.
The map is represented as lanelets~\cite{poggenhans-18}, and the simulated other vehicles are steered using the intelligent driver model (IDM)~\cite{treiber-00}.
For each CAV, \ros nodes are allocated for the AD task.
The detections are inherently created by the simulation environment.
For function offloading and other V2X technologies, e.g., track-to-track fusion or cooperative functionality, the simulation also includes \ros nodes for services running on a simulated MEC server.
For our evaluation, we assume that CAVs only offload services to the MEC server.
Lastly, we have integrated our method into our real-world experimental CAV.
The AD stack in the real-world CAV is equivalent to the one used in the simulation.

\subsection{Response Time Selection}
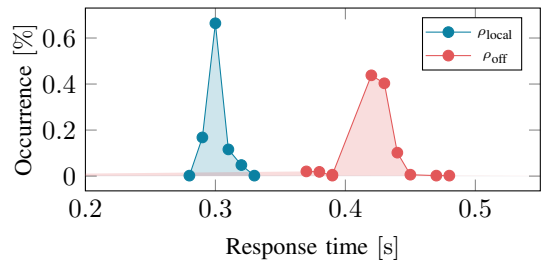
\begin{figure}[t]
\centering 
\vspace{3pt}
\begin{tikzpicture}

\def\width {\linewidth*0.88}
\def\height {4cm}
    
\pgfplotstableread[col sep=comma]{img/response_time/counts_local.csv}\local
\pgfplotstableread[col sep=comma]{img/response_time/counts_offloaded.csv}\offloaded

\def\position{(0.85,0.95)}%

\begin{axis}[
      height=\height,
      width=\width,
      xlabel={Response time $[\SI{}{\second}]$},
      xlabel style={font=\small},
      ylabel={Occurrence $[\SI{}{\percent}]$},
      ylabel style={font=\small},
      ylabel shift=-5pt,
      xmin=0.2, xmax=0.55,
      legend style={at={\position},anchor=north,legend columns=1,font=\tiny,inner sep=1pt},
    ]
    
    \addplot[name path = local, sharp plot,mark=*,muted-blue] table [x=value, y=count, col sep=comma]{\local};
    \addplot[name path = offloaded, sharp plot,mark=*,muted-red] table [x=value, y=count, col sep=comma]{\offloaded};
    
    \addplot[name path = floor, draw = none] coordinates {(0,0) (0.6,0)};
    \addplot[color=muted-blue, opacity=0.2] fill between[of = local and floor];
    \addplot[color=muted-red, opacity=0.2] fill between[of = offloaded and floor];

    \legend{$\rho_\text{local}$\vphantom{P}, $\rho_\text{off}$\vphantom{P}}
    
\end{axis}

\end{tikzpicture}

\caption{Distribution of local and offloaded response times.}

\label{fig:response}

\end{figure}

An accurate measurement of the response time is crucial for a safe application of RSS, as it is for our proposed SOFOF extension with RSS.
RSS requires a conservative estimate for response time, as an optimistic estimate would result in an insufficient safety envelope.
Furthermore, real-time AD systems suffer from tail latencies, where environmental factors, such as traffic density, increase latencies, making the most probable response time insufficient to capture worst-case behavior~\cite{Liu-2025}.
We take both factors into account when defining evaluation response times.

Since our AD stack does not allow online measurement of response time, we initially measure them offline.
This is achieved by triggering a reaction of the AD stack to a dummy object and measuring the time from object spawn to reaction.
The results of 500 simulation runs for both the local and offloaded response time are summarized in Fig.~\ref{fig:response}, with the occurrences shown in percentages.
Both plots show an obvious peak at the most probable response time.
It is also evident that the peak occurs at a lower response time for $\rho_\text{local}$ than for $\rho_\text{off}$, since the offloaded service composition requires more nodes to process the data, e.g., for V2X communication.
However, choosing only the most probable response time is insufficient for safety-critical services.
We therefore take the 0.99 quantiles of $\rho_\text{local}$ and $\rho_\text{off}$:
\begin{align}\label{eq:response}
\begin{aligned}
    \rho_\text{local} &= \SI{0.319}{\second} \\
    \rho_\text{off} &= \SI{0.450}{\second}
\end{aligned}
\end{align}
Considering an AD stack running at $\sim\SI{10}{Hz}$, a 0.99 quantile would mean that, on average, the response time is exceeded once every ten seconds; however, the next execution cycle occurs only $0.1$ seconds later and is very likely to remain within the maximum expected response time.
Thus, the $0.99$ quantile values are sufficient for our safety evaluation.

\subsection{RSS-Guarded Evaluation}
For the integration of RSS rules, we adapted the RSS definition to include a new RSS-guarded state and defined state transitions that involve a fallback preparation for faster switch back to local execution (see Fig.~\ref{fig:state-machine}).
In this evaluation, we analyze the improvement resulting from the prepared fallback with warm-standby from the RSS-guarded state compared to a direct fallback.

We consider the direct fallback to happen from full offloading to full local execution.
For this transition, we first fully activate the \ros nodes from the inactive state if the RSS-dangerous situation is detected.
Second, the proper response requires adapting the parameters of the local trajectory planner.
This is done through a \ros service call.

During fallback preparation, we already activate the \ros node at a lower update frequency, i.e., in warm-standby.
Given this activation, the parameter adjustment for the proper response can already be done.
Thus, when the RSS-dangerous situation is detected, only the update frequency of the trajectory planning node needs to be adjusted.

For our analysis, we run three scenarios in which the RSS rules are triggered 10 times each.
The average values for the switch back are given by
\begin{align}\label{eq:switch}
\begin{aligned}
    \text{normal:\ \ \ \ \ } &\rho_\text{switch}^\text{n} = \SI{95.76}{\milli\second} \\
    \text{guarded:\ \ \ \ \ } &\rho_\text{switch}^\text{g} = \SI{53.69}{\milli\second},
\end{aligned}
\end{align}

As seen in Eq.~\eqref{eq:switch}, the switch time is significantly improved by the simpler fallback enabled by the integration of the guarded state.

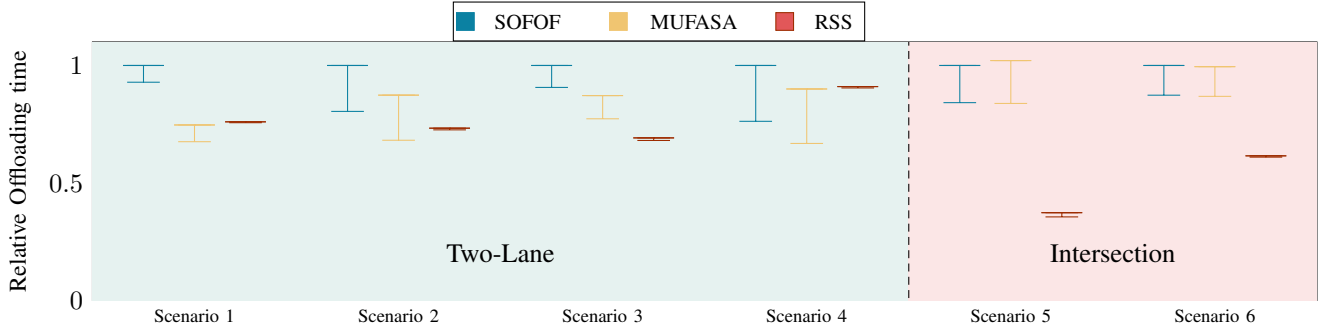
\begin{figure*}[t]
\centering
\vspace{3pt}
\begin{tikzpicture}
    \def\legendxshift{.2cm};
    \node (sofof-c) [draw=muted-blue, fill=muted-blue] {};
    \node (sofof) [right of=sofof-c, xshift=-5pt, font=\footnotesize] {SOFOF};
    \node (mufasa-c) [right of=sofof, xshift=\legendxshift, draw=muted-gold, fill=muted-gold] {};
    \node (mufasa) [right of=mufasa-c, font=\footnotesize] {MUFASA};
    \node (rss-c) [right of=mufasa, xshift=\legendxshift, draw=muted-darkred, fill=muted-red] {};
    \node (rss) [right of=rss-c, xshift=-10pt, font=\footnotesize] {RSS};

    \node[draw=black,fit=(sofof-c)(rss), inner sep=1pt] (group) {};
\end{tikzpicture}
\vspace{.1cm}
\centering    
\begin{tikzpicture}
\pgfplotsset{width=\linewidth,height=5cm,compat=1.18}
\JSONParseFromFile{\myJSONdata}{img/evaluation/processed.json}
\begin{axis}[
    boxplot/draw direction=y,
    ylabel={Relative Offloading time},
    ylabel style={font=\small},
    xtick={2, 6, 10, 14, 18, 22},
    xticklabels={{Scenario 1},{Scenario 2},{Scenario 3},{Scenario 4},{Scenario 5},{Scenario 6}},
    xticklabel style={font=\scriptsize},
    xtick pos=bottom,
    xmin=0,xmax=24,
    ymin=0,ymax=1.1
  ]

    \fill[muted-teal!15] (axis cs:0,0) rectangle (axis cs:16,1.1);
    
    \fill[muted-red!15] (axis cs:16,0) rectangle (axis cs:24,1.1);
        
    \newcounter{position}
    \setcounter{position}{1}
    \foreach \scenariona in {base,base2,base3,base4,intersection2,intersection3}  
    {
        \foreach \alg/\algcolor in {SOFOF/muted-blue,MUFASA/muted-gold,RSS/muted-darkred}{

            \edef\kmed{\scenariona.\alg[0]}
            \edef\violations{\scenariona.\alg[1]}
        
            \JSONParseValue[store in=\Vmed]{\myJSONdata}{\kmed}
            \JSONParseValue[store in=\vio]{\myJSONdata}{\violations}
        
            \edef\drawpos{\theposition}
            
            \edef\myboxplot{%
              \noexpand\addplot [
                boxplot prepared={%
                  draw position=\drawpos,
                  median=\Vmed,
                  lower quartile=\Vmed,
                  upper quartile=\Vmed,
                  lower whisker=\vio,
                  upper whisker=\Vmed,
                },
                \algcolor, fill=\algcolor!60
              ] coordinates {};%
            }
            \myboxplot

            \addtocounter{position}{1}
        }
        \addtocounter{position}{1}
    }
    \addplot +[mark=none, black] coordinates {(16, 0.01) (16, 1.1)};
    \node at (axis cs:8,0.2) {Two-Lane};
    \node at (axis cs:20,0.2) {Intersection};
\end{axis}

\end{tikzpicture}

\caption{Comparison of the average offloading times for different scenarios for SOFOF, MUFASA, and RSS, normalized to the offloading time of SOFOF for each scenario, respectively. Summarized are 4 two-lane scenarios and 2 intersection scenarios. The lower value of the plots indicates the RSS-safe offloading time; consequently, the length of the line in between the two values is the relative time when offloading was RSS-dangerous.}

\label{fig:offloading-time}

\end{figure*}

\subsection{Safety Evaluation}
Our main motivation for integrating RSS into SOFOF was to improve safety during offloaded AD service compositions.
In the following, we analyze the performance of the extended framework compared to SOFOF with a simple location-based decision-making algorithm~\cite{dehler-25}.
For an extensive evaluation, we also compare our method with MUFASA~\cite{dehler-26}, our previous state-of-the-art technique for assessing offloaded services.
Note that QoS requirements are already built into SOFOF.
Consequently, since our RSS method and MUFASA both extend SOFOF, they are inherently considered in these approaches as well.
Since we use the same QoS requirements for all three approaches, a specific QoS analysis is excluded.

In the simulation, the service orchestrator running on the CAV for all methods can select an offloaded service composition that includes an available remote trajectory planning service.
The area where offloading is possible encompasses the entire simulation environment.
For the following analysis, we set the previously determined response times from the 0.99th quantile and the determined guarded switch-back time, i.e., $\rho_\text{local} = \SI{0.319}{\second}$, $\rho_\text{off} = \SI{0.450}{\second}$, and $\rho_\text{switch}^\text{g} = \SI{0.054}{\second}$.

Figure~\ref{fig:offloading-time} summarizes the results of our simulations, showing several different scenarios where the RSS rules are applied.
In two-lane scenarios (scenarios 1-4), the other vehicles are mapped to the lane of the ego vehicle, whereas in intersection scenarios (scenarios 5-6), the environment is more complex, with multiple lane geometries.
We ran each scenario in 10 closed-loop simulations for each approach, i.e., SOFOF, MUFASA, and RSS.

Each plot shows two values: the upper bars show the median relative offloading time, normalized to the offloading time of SOFOF; the lower bars show the relative time during which offloading was RSS-safe or RSS-guarded.
Comparing the median values, the offloading efficiency, which is proportional to the offloading duration, is reduced when using safety-based methods for offloading decision-making and fallback, with MUFASA being more efficient in the majority of the scenarios compared to RSS.
However, when looking at the intervals between the two bars, as shown by the connecting line, an important advantage of the RSS-based method becomes apparent.
The interval, i.e., the length of the line, directly shows the duration for which the situation was RSS-dangerous while function offloading was activated.
It can be seen that, when using the RSS-based method, the scenarios are significantly safer when the RSS rules are used as a safety analysis metric.

In the intersection scenarios, efficiency is much lower than in the two-lane scenarios.
This is a consequence of more complex scenarios, which increase the likelihood of RSS-dangerous situations.
Nevertheless, the safety of the scenario is still improved through the safety assessment of the offloaded service, again seen by the interval.

Summarizing over all simulations shown in Fig.~\ref{fig:offloading-time}, the average count for how often the situation was RSS-guarded before it changed to RSS-dangerous was $15.5$.
This value indicates that the vehicle keeps the offloading active for an average of $15.5$ time steps, providing enough time for the fallback preparation.
This value also reaffirms the difference seen in the switch-back times in Eq.~\eqref{eq:switch}.

The lower relative offloading times, combined with the increased safety in the scenarios, demonstrate the described trade-off between safety and efficiency.
These results are expected yet significant, given that RSS is a well-recognized method for safety analysis.

\subsection{Real-World Deployment}

To demonstrate the significance of our proposed method beyond simulation, we have also integrated it into our real-world experimental CAV, which has an equivalent setup, i.e., a modular AD stack with \ros.
The function offloading functionality is provided by a MEC server running the additional \ros nodes for offloading, and a dedicated 5G network enables V2X communication. The V2X setup is similar to that described in~\cite{buchholz-26}.

\begin{figure}[t]
\centering
\includegraphics[width=.95\linewidth]{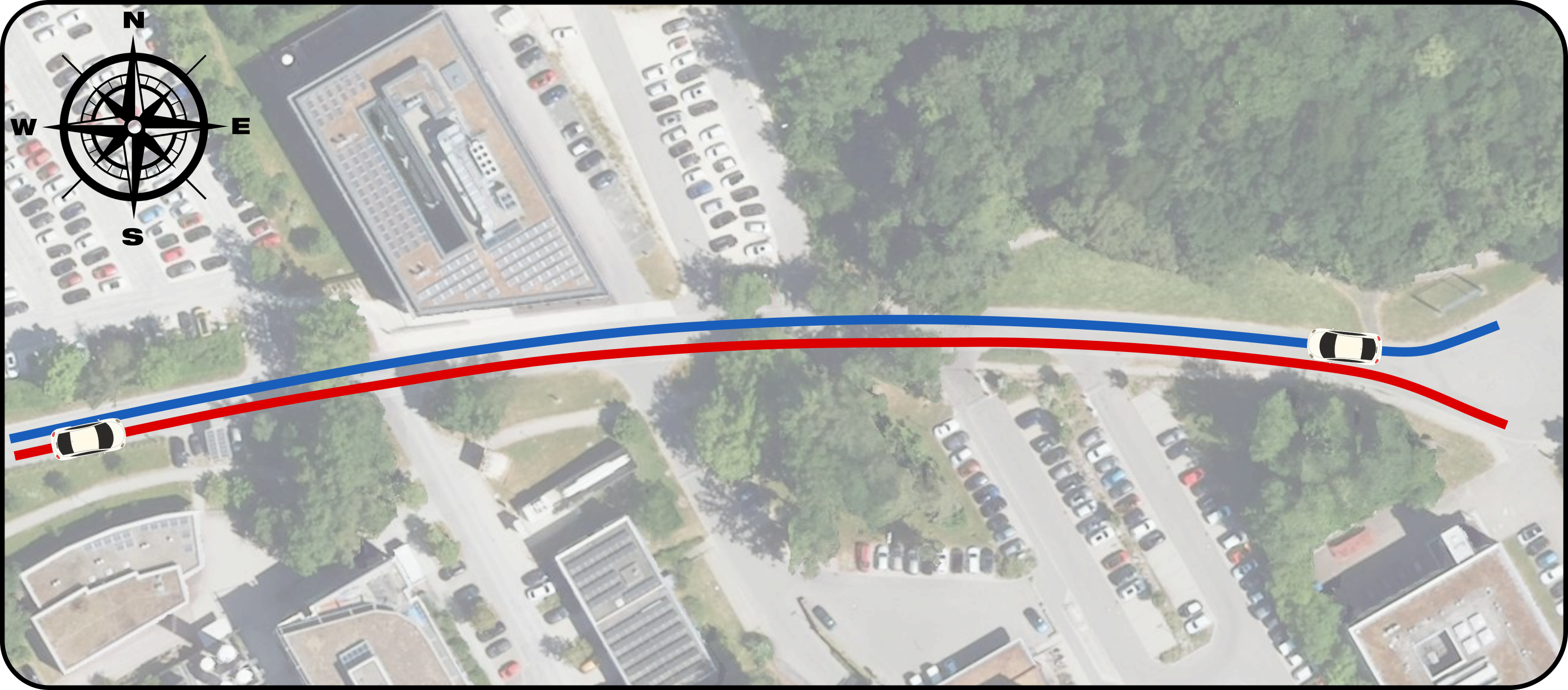}
\caption{Street layout and routes of the real-world test drives at Helmholtzstraße, Ulm. Satellite picture from~\cite{Map-Tiles}.}
\label{fig:routes}
\end{figure}

For our real-world evaluation, the CAV drives with offloaded trajectory planning, while a second vehicle intentionally triggers the RSS-guarded and RSS-dangerous scenarios.
Since we operate in real-world traffic, the CAV may also react to other vehicles.
Unlike the simulation, in the real-world AD stack, we have to adjust response times to account for the additional time required by the sensors and object detection.
To create critical scenarios more easily, we added a constant offset of $\SI{0.3}{\second}$ to the response times, as longer response times lead to greater safety distances.\footnote{Even though the response times then do not represent the actual response time of the real-world CAV, with the difference of $\rho_\text{local}$ and $\rho_\text{off}$ being consistent, they are adequate for proof-of-concept.}
Thus, for the real-world evaluation, we have $\rho_\text{local}= \SI{0.789}{\second}$ and $\rho_\text{off} = \SI{0.87}{\second}$.
With this setup, we conducted 8 real-world test drives on two routes shown in Fig.~\ref{fig:routes}, both two-lane scenarios.

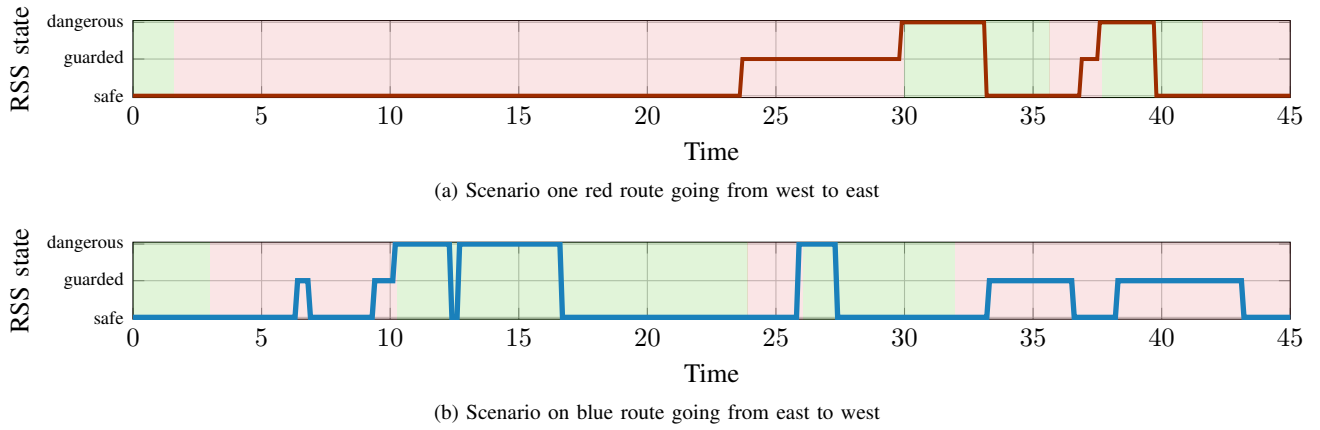
\begin{figure*}[t]

\centering
\def\heightreal{2.6cm}
\subfloat[Scenario one red route going from west to east]{%
\label{fig:real_world_one}
\begin{tikzpicture}
    \pgfplotsset{width=.95\linewidth,height=\heightreal,compat=1.18}
    \pgfplotstableread[col sep=comma]{img/evaluation/real_world/20260507_12_25_20.csv}\one
    
    \begin{axis}[
        xlabel={Time},
        ylabel={RSS state},
        xmin=0.0, xmax=45.0,
        xtick distance=5,
        ymin=-0.05, ymax=2.05,
        ytick={0, 1, 2}, yticklabels={{safe}, {guarded}, {dangerous}},
        yticklabel style={font=\scriptsize},
        grid=major
    ]
    
    \fill[muted-green, opacity=0.15] (axis cs:-0.1,-0.1) rectangle (axis cs:1.6,3.1);
    \fill[muted-red, opacity=0.15] (axis cs:1.6,-0.1) rectangle (axis cs:29.98,3.1);
    \fill[muted-green, opacity=0.15] (axis cs:29.98,-0.1) rectangle (axis cs:35.62,3.1);
    \fill[muted-red, opacity=0.15] (axis cs:35.62,-0.1) rectangle (axis cs:37.67,3.1);
    \fill[muted-green, opacity=0.15] (axis cs:37.67,-0.1) rectangle (axis cs:41.59,3.1);
    \fill[muted-red, opacity=0.15] (axis cs:41.59,-0.1) rectangle (axis cs:45.1,3.1);    
    
    \addplot[no marks, muted-darkred, line width=1.5pt] table [col sep=comma, x=time, y=rss] {\one};
    
    \end{axis}
    
\end{tikzpicture}

}%
\vspace{.2cm}
\subfloat[Scenario on blue route going from east to west]{%
\label{fig:real_world_two}
\begin{tikzpicture}
    \pgfplotsset{width=.95\linewidth,height=\heightreal,compat=1.18}
    \pgfplotstableread[col sep=comma]{img/evaluation/real_world/20260507_12_36_04.csv}\two
    
    \begin{axis}[
        xlabel={Time},
        ylabel={RSS state},
        xmin=0.0, xmax=45.0,
        xtick distance=5,
        ymin=-0.05, ymax=2.05,
        ytick={0, 1, 2}, yticklabels={{safe}, {guarded}, {dangerous}},
        yticklabel style={font=\scriptsize},
        grid=major
    ]
    
    \fill[muted-green, opacity=0.15] (axis cs:-0.1,-0.1) rectangle (axis cs:3,3.1);
    \fill[muted-red, opacity=0.15] (axis cs:3,-0.1) rectangle (axis cs:10.27,3.1);
    \fill[muted-green, opacity=0.15] (axis cs:10.27,-0.1) rectangle (axis cs:23.9,3.1);
    \fill[muted-red, opacity=0.15] (axis cs:23.9,-0.1) rectangle (axis cs:26.05,3.1);
    \fill[muted-green, opacity=0.15] (axis cs:26.05,-0.1) rectangle (axis cs:31.96,3.1);
    \fill[muted-red, opacity=0.15] (axis cs:31.96,-0.1) rectangle (axis cs:45.1,3.1);
    \addplot[no marks, custom-blue, line width=2pt] table [col sep=comma, x=time, y=rss] {\two};
    
    \end{axis}
    
\end{tikzpicture}

}%

\caption{RSS state transitions of two representative real-world test drives. In the background, the \ros lifecycle state of the local trajectory planning service is visualized (green: active, red: inactive).}

\label{fig:real-world}

\end{figure*}

In Fig.~\ref{fig:real-world}, we present two representative drives, visualizing the RSS states over time with the \ros lifecycle node state color-coded in the background: green indicates that local trajectory planning is active, while red indicates it is inactive.\footnote{Inactive trajectory planning is equivalent to using an offloaded trajectory planning service.}

In Fig.~\ref{fig:real-world}(a), the red line shows the expected behavior regarding the RSS state transitions.
After around $\SI{2}{\second}$, the offloading of the trajectory planning service started, which is visible in the change of the background color from green to red.
At this point, the situation is still RSS-safe.
Starting at \mbox{$\sim\SI{24}{\second}$}, the RSS state changes from RSS-safe to RSS-guarded.
At this point, a vehicle that was normally present in the traffic was detected, which caused the transition.
In combination, the SOFOF and RSS modules began preparing the fallback at this time. 
Then, at $\sim\SI{30}{\second}$, the safety distance was so small that, even with the local response time, the situation was RSS-dangerous, as seen in the transition from RSS-guarded to RSS-dangerous.
At this point, the proper response was triggered, and the local trajectory planning service was activated immediately, as indicated by the switch to the green background.
During local execution, the state recovers to RSS-safe at $\sim\SI{33.5}{\second}$.
Offloading is then again activated at $\sim\SI{35.5}{\second}$.
Shortly after, our second vehicle deliberately created the dangerous situation at $\sim\SI{37}{\second}$, triggering a similar state transition cycle from RSS-safe to RSS-guarded to RSS-dangerous.
Equivalently, the local trajectory planning service was activated, and the state recovered to the RSS-safe state again after some time.

In Fig.~\ref{fig:real-world}(b), the blue line shows a busier scenario, where several other vehicles present in the traffic caused multiple state transitions.
Similar to the argumentation before, our second vehicle causes the transition cycle from RSS-safe to RSS-guarded to RSS-dangerous at $\sim\SI{9.5}{\second}$.
However, one of the main benefits of extending RSS with the RSS-guarded state becomes apparent in this scenario.
This is especially seen at each of the times $\sim\SI{6}{\second}$, $\sim\SI{33}{\second}$, and $\sim\SI{38}{\second}$, where the state changes from RSS-safe to RSS-guarded and back to RSS-safe.
Most importantly, the local trajectory planning service is not activated, as shown by the background color.
Without the extension of RSS by the state RSS-guarded, the framework would classify these scenarios directly as RSS-dangerous and activate the local trajectory planning service.
At the times $\sim\SI{12.5}{\second}$ and $\sim\SI{25.5}{\second}$, a direct switch from RSS-safe to dangerous is visible, skipping the RSS-guarded state since the situation is too critical with respect to both $\rho_\text{off}$ and $\rho_\text{local} + \rho_\text{switch}$.

Lastly, we summarize the number of occurrences of important state transitions during the 8 test drives.
During the real-world evaluation drives, a switch back to local execution was prevented 7 times (RSS-safe $\rightarrow$ RSS-guarded $\rightarrow$ RSS-safe), the fallback preparation helped to improve the transition to local execution 6 times (RSS-safe $\rightarrow$ RSS-guarded $\rightarrow$ RSS-dangerous), while the undesirable transition (RSS-safe $\rightarrow$ RSS-dangerous) only occurred 4 times.

In summary, our method maintains safety during offloading when necessary, while focusing on efficiency when the switch back is not yet required.

\section{Conclusion and Future Work}\label{ch:conclusion}
In this work, we have presented an extension of RSS to handle different response times of local and offloaded service compositions for distributed SOAs of CAVs.
We described a method for using the RSS extension to support safer offloading decision-making and a fallback preparation approach for a faster, thus safer, switch back to local execution.
In our evaluation, we critically analyzed our method against two state-of-the-art offloading frameworks, demonstrating safety improvements at a bearable expense of efficiency.

In our future work, we want to further enhance safety for function offloading by including predictive QoS measures that take into account not only traffic scenarios but also network conditions.
Furthermore, we want to analyze the application of function offloading and the associated safety considerations across different AD approaches, e.g., by deploying end-to-end AD networks on MEC and cloud servers.

\balance
\bibliographystyle{utility/IEEEtran-et-al}
\bibliography{biblio}
	
\end{document}